\documentclass[sigconf]{acmart}

\copyrightyear{2026}
\acmYear{2026}
\setcopyright{cc}
\setcctype{by}
\acmConference[WWW '26]{Proceedings of the ACM Web Conference 2026}{April 13--17, 2026}{Dubai, United Arab Emirates}
\acmBooktitle{Proceedings of the ACM Web Conference 2026 (WWW '26), April 13--17, 2026, Dubai, United Arab Emirates}
\acmPrice{}
\acmDOI{10.1145/3774904.3793019}
\acmISBN{979-8-4007-2307-0/2026/04}

\usepackage{amsthm}
\newtheorem{myDef}{Definition}
\usepackage{booktabs} 
\usepackage{multirow}
\usepackage{colortbl}
\usepackage{amsmath}
\usepackage{listings}
\usepackage{mdframed}
\usepackage{tcolorbox}
\usepackage{dsfont}
\usepackage{bbm}

\begin{document}

\title{Intelli-Planner: Towards Customized Urban Planning via Large Language Model Empowered Reinforcement Learning}

\author{Xixian Yong}
\affiliation{%
  \institution{Gaoling School of Artificial Intelligence\\Renmin University of China}
  \city{Beijing}
  \country{China}
}
\email{xixianyong@ruc.edu.cn}

\author{Peilin Sun}
\affiliation{%
  \institution{Gaoling School of Artificial Intelligence\\Renmin University of China}
  \city{Beijing}
  \country{China}
}
\email{sunpeilin63@ruc.edu.cn}

\author{Zihe Wang}
\affiliation{%
  \institution{Gaoling School of Artificial Intelligence\\Renmin University of China}
  \city{Beijing}
  \country{China}
}
\email{wang.zihe@ruc.edu.cn}

\author{Xiao Zhou}
\authornote{Corresponding author.}
\authornote{Also with Beijing Key Laboratory of Research on Large Models and Intelligent Governance.}
\authornote{Also with Engineering Research Center of Next-Generation Intelligent Search and Recommendation, MOE.}
\affiliation{%
  \institution{Gaoling School of Artificial Intelligence\\Renmin University of China}
  \city{Beijing}
  \country{China}
}
\email{xiaozhou@ruc.edu.cn}


\begin{abstract}
Effective urban planning is crucial for enhancing residents' quality of life and ensuring societal stability, playing a pivotal role in the sustainable development of cities. Current planning methods heavily rely on human experts, which are time-consuming and labor-intensive, or utilize deep learning algorithms, often limiting stakeholder involvement. To bridge these gaps, we propose Intelli-Planner, a novel framework integrating Deep Reinforcement Learning (DRL) with large language models (LLMs) to facilitate participatory and customized planning scheme generation. Intelli-Planner utilizes demographic, geographic data, and planning preferences to determine high-level planning requirements and demands for each functional type. During training, a knowledge enhancement module is employed to enhance the decision-making capability of the policy network. Additionally, we establish a multi-dimensional evaluation system and employ LLM-based stakeholders for satisfaction scoring. Experimental validation across diverse urban settings shows that Intelli-Planner surpasses traditional baselines and achieves comparable performance to state-of-the-art DRL-based methods in objective metrics, while enhancing stakeholder satisfaction and convergence speed. These findings underscore the effectiveness and superiority of our framework, highlighting the potential for integrating the latest advancements in LLMs with DRL approaches to revolutionize tasks related to functional areas planning. Code and data are available at \url{https://github.com/chicosirius/Intelli-Planner}.
\end{abstract}



\maketitle

\section{Introduction}
As urbanization advances, urban communities become central hubs for various human activities like living, working, education, healthcare, and commerce~\citep{haapio2012towards,li2025curegraph}. Designing effective schemes for functional area planning to organize these vital urban spaces not only enhances residents' daily experiences but also plays a crucial role in shaping a city's culture, fostering economic development, and ensuring sustainability~\citep{bai2020evolution,yong2024musecl,li2025fap}. Despite its importance, planning tasks have traditionally relied on human expertise. Considering the complexity of urban systems, planning schemes require multidisciplinary collaboration, which is time-consuming and labor-intensive. For example, completing a planning project for Greenwich Village in Manhattan may require a team of approximately 30 people, working continuously for about six months to one year~\citep{harris2003around}.

With the evolution of information technology, tools such as Geographic Information Systems (GIS)~\citep{maguire1991overview}, have emerged to alleviate the burden. Furthermore, recent advances in deep learning techniques, particularly Generative Adversarial Network (GAN)-based methods, have paved the way for automated scheme generation. They can be divided into two categories: a) generate schemes based on the simulation of urban development patterns without detailed land use configurations~\citep{isola2017image, albert2019spatial, zhang2022metrogan}, and b) more fine-grained generated schemes using green rate levels as human input for planning~\citep{wang2023human, wang2020reimagining}. Recently, a spatial optimization method based on deep reinforcement learning (DRL) has been proposed to further advance automated urban planning~\citep{zheng2023spatial}.

While these methods have demonstrated promising results, they neglect stakeholder preferences, thereby marginalizing diverse voices~\citep{forester1999deliberative}. Traditional participatory planning typically involves reconciling the interests of various groups through surveys and interviews, a process demanding substantial time and human resources~\citep{hong2012land}. The emergence of large language models (LLMs) provides a promising solution, leveraging their remarkable ability to comprehend external inputs and generate human-like feedback~\citep{brown2020language, wang2024survey, yong2025motivebench}. This capability offers valuable insights that can enhance urban planning practices.

Another critical issue concerns the evaluation of planning schemes. Recent studies have begun to focus on service and ecology~\citep{zheng2023spatial}, or sustainability, diversity, and equity among stakeholders~\citep{qian2023ai}. However, most existing methods rely heavily on subjective qualitative measures. Hence, proposing a comprehensive evaluation approach for planning schemes is increasingly necessary.

To tackle these challenges, we explore integrating recent advancements in LLMs to revolutionize functional area planning tasks and enhance participation in the planning process. However, relying solely on LLMs for planning may lead to unreliable results due to issues like hallucination~\citep{ji2023survey}. Moreover, the speed and cost of API calls limit the scalability of generating planning schemes for larger communities. Existing literature has shown that combining LLMs with DRL can significantly enhance downstream task effectiveness. For instance, LLMs play a crucial role in generating state transitions or reward functions within the environment~\citep{xie2023text2reward} and can generate actions or provide higher-level guidance for agents~\citep{carta2023grounding, brohan2023can}.

Thus, in this study, we introduce \textit{Intelli-Planner}, an innovative urban planning framework that leverages LLMs' domain knowledge and role-playing strengths alongside DRL's optimization capabilities to generate and update planning schemes. Our contributions include:

\begin{itemize}
    \item We pioneer the integration of LLM into DRL, leveraging LLM's strengths in domain knowledge and role-playing, along with DRL's advantages in optimization, to accomplish the generation and update of urban planning schemes.
    \item We design a five-dimensional scheme evaluation system, which includes objective indicators such as \textit{service}, \textit{ecology}, \textit{economy}, and \textit{equity}, along with \textit{satisfaction} scores from various LLM-based city stakeholders.
    \item Experiments conducted in three communities with different styles and demands show that Intelli-Planner exhibits universality and superiority, making it applicable to a wide range of practical planning scenarios.
\end{itemize}

\begin{figure*}[t]
\centering
\includegraphics[width=1\textwidth]{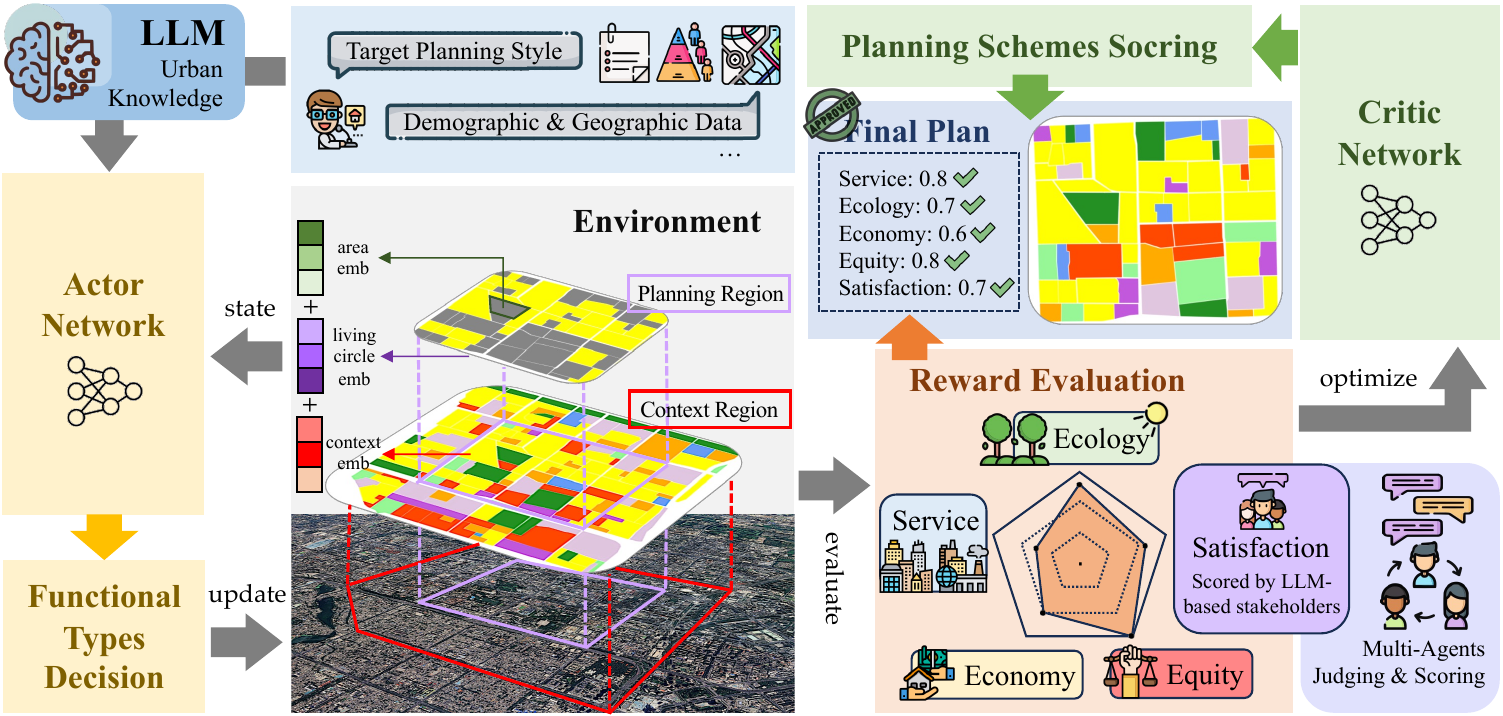}
\caption{The overall architecture of Intelli-Planner. It follows an Actor-Critic architecture, including an Actor network that makes decisions on functional area types, and a Critic network that is continuously optimized through evaluation system.}
\label{framework}
\end{figure*}

\section{Related Work}
\subsubsection{Urban Functional Areas Planning.}
Land functional types allocation is a crucial aspect of spatial optimization research, extensively studied~\cite{liao2022land,wang2021multi}. Traditional planning relies on urban designers' intuition, emphasizing aesthetics and functionality~\cite{levy2016contemporary}. Then some techniques like multi-objective optimization~\cite{lan2015multi} and swarm intelligence algorithms~\cite{gao2017improved} have been utilized, aiming to achieve various objectives. With the development of deep learning, models like GAN~\cite{wang2020reimagining} and Variational Autoencoder (VAE)~\cite{wang2023towards} show promise in urban layout generation. Specifically, some simulate urban development patterns using vanilla GAN~\cite{albert2018modeling} or conditional GAN of Pix2Pix~\cite{isola2017image,albert2019spatial, zhang2022metrogan}, and more detailed approaches, such as those by~\citeauthor{shen2020machine}~\shortcite{shen2020machine}, generate images with building shapes, while~\citeauthor{wang2022human}~\shortcite{wang2022human}
incorporate land-use configurations, using green rate levels for planning. Recently, DRL has emerged as a powerful approach for spatial optimization~\cite{zheng2023spatial}. However, these approaches often overlook collective decision-making and stakeholder interdependence. While efforts have been made to include stakeholders~\cite{qian2023ai,abolhasani2022collective}, participatory planning remains an urgent issue.

\subsubsection{Large Language Model Based Agents.}
Humans have long pursued artificial intelligence that matches or surpasses human capabilities~\cite{xi2023rise}. Recently, LLMs have shown remarkable potential in achieving human-level capabilities through vast web knowledge, offering hope for general AI agents~\cite{wang2023survey}. Researchers are using LLMs to build AI agents, exploring single-agent deployment~\cite{gur2023real,li2023hitchhiker,wang2023voyager}, multi-agent interaction~\cite{hao2023chatllm,wu2023autogen}, and human-agent interaction~\cite{hasan2023sapien}. Studies also investigate agent personality~\cite{binz2023using,caron2022identifying}, behavior~\cite{shinn2023reflexion,chen2023agentverse}, collective intelligence~\cite{liu2023training,chan2023chateval}, and social phenomena~\cite{park2023generative}. These advancements have applications in political science~\cite{argyle2023out}, jurisprudence~\cite{cui2023chatlaw}, civil engineering~\cite{mehta2023improving}, robotics~\cite{nottingham2023embodied}, and more.

\section{Problem Statement}
To simplify and concretize the meaning of urban planning, we define it as a customized urban reconstruction problem:
\begin{myDef}
\textbf{(Customized Urban Reconstruction.)} \rm Assuming there is a region $\mathcal{R}$, with demographic data represented by $\mathcal{D}$, such as age distribution, educational attainment, income and consumption levels, etc. The areas $r_i \in \mathcal{R}$ exhibit geographic attributes denoted as $g_i \in \mathcal{G}$, encompassing the functional type, location, area, perimeter, and compactness, among others. Given a planning objective $\mathcal{O}$ in the form of natural language, the task involves determining the functional types of each designated area, ultimately resulting in a scheme $\mathcal{P}$ that simultaneously achieves high evaluation metrics and satisfies the interests of various stakeholders.
\end{myDef}

\section{Methodology}
\subsection{Framework Overview}
To integrate the world knowledge capabilities of LLM with the search proficiency of reinforcement learning, we propose a participatory and customized urban planning framework, as depicted in Figure \ref{framework}. This Actor-Critic-based framework uses LLM for formulating planning objectives, enhances the policy network with single-agent guidance, and achieves stakeholder satisfaction through multi-agent scoring. We also introduce a five-dimensional evaluation system for a comprehensive assessment of the final schemes.

\subsection{Planning Objectives Formulation}
\label{demands}
\subsubsection{Initial Settings.}
In the initial phase of city planning for a given region, we segment the target area into multiple polygonal zones based on the available road network and building distribution data from OpenStreetMap\footnote{https://www.openstreetmap.org/} (OSM). Since the land use tag in OSM might not be consistently available for each small area, we utilize Points of Interest (POI) and building types data on OSM to determine the current functional type for each polygonal area. These functional types include Residential (RS), Business (BU), Office (OF), Recreation (RC), School (SC), Hospital (HO), Park (PA), and Open Space (OP):
\begin{equation}
    {\rm Func} = \{{\rm RS}, {\rm BU}, {\rm OF}, {\rm RC}, {\rm SC}, {\rm HO}, {\rm PA}, {\rm OP}\}.
\end{equation}
After obtaining the actual functional distribution map for the region, we retain only residential areas, masking other types as areas that need reconstruction.

\subsubsection{Customized Demands \& Planning Requirements.}
Prior studies have indicated that LLMs demonstrate remarkable natural language understanding and decision-making abilities across various domains~\citep{brown2020language, xi2023rise, wang2023survey}. Through extensive training on vast corpora, general models like GPT-3.5 embody rich urban planning knowledge. Therefore, an intuitive approach is to leverage their capabilities for high-level planning objectives formulation.

\begin{figure}[htbp]
\centering
\includegraphics[width=1\columnwidth]{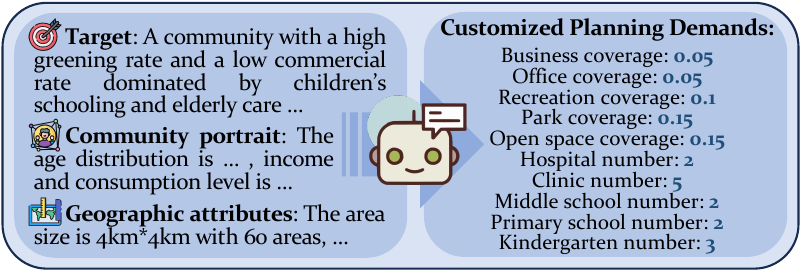}
\caption{Formulate planning objectives according to the target style and the area's fundamental information.}
\label{llm1}
\end{figure}

Specifically, we begin by gathering demographic and geographical data for the planning region under consideration. These include age distribution, income levels, consumption patterns for both residential and working populations, as well as original functional types, areas, and shapes of individual land parcels. These inputs, along with our defined target planning style, are then fed into the LLM, which provides expected coverage rates and quantities for each functional type. The schematic diagram is shown in Figure~\ref{llm1}.

\subsection{Reconstruction Schemes Decision}
\subsubsection{Markov Decision Process.}
Given the dynamic interactivity of urban planning, we conceptualize it as a sequential Markov decision process (MDP)~\citep{bellman1957markovian}, where the planner (\textit{agent}) interacts with urban areas (\textit{environment}). At each step, the planner observes the current scheme (\textit{state}), selects an area's functional type (\textit{action}), and alters land configuration (\textit{transition}). The planner receives feedback (\textit{reward}) and iteratively adjusts its strategy until convergence. It's represented by the four-tuple $<S, A, P, R>$.

(1) \textbf{State space $S$}. To accurately characterize the planning state of the urban, we choose to describe it on three levels: the geographic attributes of each area, the distribution of functional zones within community living circles, and the planning statistics data for the entire region. For area $r_i$, its geographic attributes feature is:
\begin{equation}
    g_i = [{\rm Type}_i, {\rm Coor}_i, {\rm Area}_i, {\rm Peri}_i, {\rm Comp}_i],
\end{equation}
where the five features are type, coordinate location, area, perimeter, and compactness of $r_i$. The geographic attributes constrain the functional types that can be constructed in this area. Furthermore, the community living circle distribution feature of area $r_i$ is:

\begin{equation}
    c_i = [{\rm Ratio}_1^{i}, \cdots, {\rm Ratio}_{n_{\rm Func}}^{i}, {\rm Count}_1^{i}, \cdots, {\rm Count}_{n_{\rm Func}}^{i}]
\end{equation}
where ${\rm Ratio}_j^{i}$ and ${\rm Count}_j^{i}$ represent the coverage and number of the $j^{\rm th}$ functional type in the living circle of area $r_i$ respectively. Similarly, the statistical feature for the entire region is:
\begin{equation}
    e = [{\rm Ratio}_1, \cdots, {\rm Ratio}_{n_{\rm Func}}, {\rm Count}_1, \cdots, {\rm Count}_{n_{\rm Func}}]
\end{equation}
where ${\rm Ratio}_j$ and ${\rm Count}_j$ represent the coverage and number of the $j^{\rm th}$ functional type in the whole region we considered.

(2) \textbf{Action space $A$}. Traditional urban planning typically addresses three key questions: What to build? Where to build? How much area to allocate? This results in an excessively large action space, making training challenging~\citep{zheng2023spatial}. To simplify the decision-making process, we predetermine the planning sequence for the areas to be planned. In the urban reconstruction problem under consideration, the area of each plot is fixed. Therefore, we only need to address the question of "what to build." In addition, to make the planning more reasonable, we predefined certain rules that determine the rational establishment of functional types based on the geographic attributes $g_i$ of the area $r_i (i = 1, \cdots, N)$.

(3) \textbf{Transition model $P$}. Based on the Markov property, the state at a given moment depends only on the state at the previous moment and is not influenced by past states. When the current planning state $s$ and action $a$ are determined, we can obtain the probability of transitioning to the next state:
\begin{equation}
    P_a(s, s') = \mathbb{P}(S_{t+1} = s' \mid S_t=s, A_t=a).
\end{equation}
As a result, the state of the region environment changes with each action we take, and the choice of action depends on the current state of the environment.

\subsubsection{Policy \& Value Networks.}
Then we introduce the policy network enhanced by LLM and the value network for evaluating the planning schemes.

\begin{figure}[ht]
\centering
\includegraphics[width=1\columnwidth]{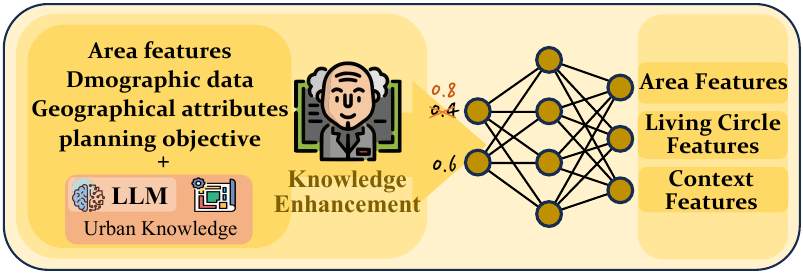}
\caption{Knowledge enhancement of the decisions.}
\label{llm2}
\end{figure}

(1) \textbf{Policy network $\pi$}. After pre-training on a large corpus, LLMs encapsulate a broad range of foundational knowledge across multiple domains, including urban planning. Therefore, we contemplate leveraging its insights to supplement priors in the action selection of the policy network. Given the demographic data $\mathcal{D}$ and geographical attributes $\mathcal{G}$ of the considered area $r_i$, along with our planning objective $\mathcal{O}$, LLM can provide a recommended set of land use types:
\begin{equation}
    {\rm RecSet}_i = {\rm Planner}(r_i, \mathcal{D}, \mathcal{G}, \mathcal{O}),
\end{equation}
where ${\rm Planner}$ is an LLM-based agent. Simultaneously, the policy network generates action probability distribution based on the geographical attribute $g_i$, the living circle distribution feature $c_i$, and the region statistical feature $e$:
\begin{equation}
    \pi_{\theta}(A \mid s_i) = {\rm FFN}_1([g_i \parallel c_i \parallel e]),
\end{equation}
where ${\rm FFN}$ is a feed-forward neural network. We enhance the probabilities of actions in the recommended set ${\rm RecTypes}_i$ for area $r_i$:
\begin{equation}
    \mathbb{P}(a \mid s_i) =
    \begin{cases}
 \pi_{\theta}(a \mid s_i) & a \notin {\rm RecSet}_i\\
  \lambda \cdot \pi_{\theta}(a \mid s_i) & a \in {\rm RecSet}_i
    \end{cases},
\end{equation}
\begin{equation}
    \pi'_{\theta}(A \mid s_i) = {\rm Softmax}(\mathbb{P}(A \mid s_i)),
\end{equation}
where $\lambda$ is a hyper-parameter representing the strength of enhancement, and $\pi'_{\theta}(A \mid s_i)$ represents the action probability distribution after enhancement as shown in Figure~\ref{llm2}.

(2) \textbf{Value network $v$}. Next, we establish a value network to quantitatively evaluate the planning performance. Specifically, here we consider the statistical features of the entire region $e$ as the input to assess the quality of the planning scheme, and obtain the final evaluation through a feed-forward network:
\begin{equation}
    \hat{v} = {\rm FFN}_2(e),
\end{equation}
where $\hat{v}$ is the evaluation of the current planning condition.

\subsection{Reward Design}
\label{rewarddesign}
To assess the effectiveness of the generated plans, we design a five-dimensional evaluation framework, comprising objective indicators of \textit{service}, \textit{ecology}, \textit{economy}, \textit{equity}, and subjective \textit{satisfaction} of various LLM-based city stakeholders.

\subsubsection{Service.}
Nowadays, the concept of the 15-minute living circle has been increasingly mentioned. Its objective is to enable residents of urban communities to access essential services for living, working, shopping, healthcare, education, and entertainment within a 15-minute walking or cycling distance~\citep{moreno2021introducing}. Specifically, we consider service-oriented functional types that encompass business, office, recreation, hospital, and school. Formally, upon obtaining a planning scheme, the calculation of service reward is:
\begin{equation}
    d_{\rm ser}(i,j) = \min \{ {\rm Dis}({\rm RS}_i, {\rm ser}_j^{1, \cdots, n_{{\rm ser}_j}}) \},
\end{equation}
\begin{equation}
    r_{\rm ser} = \frac{1}{n_{\rm RS}} \sum_{i=1}^{n_{\rm RS}} \frac{1}{n_{\rm ser}} \sum_{j=1}^{n_{\rm ser}} \mathds{1} [d_{\rm ser}(i,j) \le d_{c}],
\end{equation}
where ${\rm Dis}(\cdot)$ is used to calculate the Euclidean distance, and $d_{\rm ser}(i,j)$ represents the minimum distance between the $i^{\rm th}$ residential area and the $j^{\rm th}$ service-oriented functional type area. Ultimately, our aim is to make the service areas around each residential area as comprehensive as possible.

\subsubsection{Ecology.}
In urban planning, ecological development contributes to the health of residents and the sustainability of communities. Therefore, we establish an ecological reward metric, incorporating the construction of parks and open spaces into consideration. The specific calculation is as follows:
\begin{equation}
    {\rm Cov}_{{\rm ecl}_i} = \frac{ \sum_{j=1}^{N} {\rm Area}_j \cdot \mathds{1} [{\rm Type}_j \in {\rm ecl}_i] }{ \sum_{j=1}^{N} {\rm Area}_j },
\end{equation}
\begin{equation}
    r_{\rm ecl} = \frac{1}{n_{\rm ecl}} \sum_{i=1}^{n_{\rm ecl}} \min \{ \frac{{\rm Cov}_{{\rm ecl}_i}}{{\rm Dem}_{{\rm ecl}_i}}, 1\},
\end{equation}
where ${\rm Cov}_{{\rm ecl}_i}$ and ${\rm Dem}_{{\rm ecl}_i}$ represent the actual and expected area coverage ratios, respectively, for the $i^{\rm th}$ ecological land type. As the actual ratios approach our expectations, the reward increases, encouraging the agent to optimize towards the demands we set.

\subsubsection{Economy.}
Adequate commercial and entertainment areas can enrich residents' material and spiritual life, promote community economic development, and office areas can provide employment opportunities~\cite{luazuaroiu2020sustainability}. Hence, we establish an economic evaluation metric to assess the construction of business, recreation, and office areas in the planning. Considering that the economy should be human-centric and serve the residents in the community, we aim for these facilities to be located within the 15-minute living circle of the residential areas. The metric is calculated as follows:
\begin{equation}
    {\rm RLC} = \bigcup_{i=1}^{n_{\rm RS}} {\rm LivCir}({\rm RS_i}, d_c)
\end{equation}
\begin{equation}
    {\rm Cov}_{{\rm ecn}_i} = \frac{ \sum_{j=1}^{N} {\rm Area}_j \cdot \mathbbm{1} [{\rm Type}_j \in {\rm ecn}_j \wedge {\rm Coor}_i \in {\rm RLC}]}{ \sum_{j=1}^{N} {\rm Area}_j }
\end{equation}
\begin{equation}
    r_{\rm ecn} = \frac{1}{n_{\rm ecn}} \sum_{i=1}^{n_{\rm ecn}} \min \{ \frac{{\rm Cov}_{{\rm ecn}_i}}{{\rm Dem}_{{\rm ecn}_i}}, 1\}
\end{equation}
where ${\rm LivCir}({\rm RS_i}, d_c)$ represents the living circle area with $\rm RS_i$ as the center and $d_c$ as the radius, and $\rm RLC$ represents the union area of all residential living circles. In other words, we consider economic facilities within the living circle of residential areas for effective planning, hoping that the agent can plan ${\rm Cov}_{{\rm ecn}_i}$ according to the demands set by ${\rm Dem}_{{\rm ecn}_i}$.

\subsubsection{Equity.}
Ensuring the well-being of vulnerable populations is a crucial aspect of urban planning. With a specific focus on two groups—children and the elderly—we prioritize addressing equity issues related to schools and hospitals in the planning process. Excessive construction of these facilities can lead to resource wastage, while inadequate construction may result in resource constraints. Moreover, the current trend emphasizes proximity in education and healthcare, underscoring the significance of their locations~\citep{pineda2020disability, bryan2020fostering}. To ensure equitable access to education and healthcare in each community, we establish a dedicated equity metric:
\begin{equation}
    d_{\rm equ}(i,j) = \min \{ {\rm Dis}({\rm RS}_i, {\rm equ}_j^{1, \cdots, n_{{\rm equ}_j}}) \},
\end{equation}
\begin{equation}
    r_{\rm equ} = \frac{1}{n_{\rm equ}} \sum_{j=1}^{n_{\rm equ}}
 e^{\, \alpha ( \max\limits_{i} \{d_{\rm equ}(i,j)\} - \min\limits_{i} \{d_{\rm equ}(i,j)\} )},
\end{equation}
where $d_{\rm equ}(i,j)$ represents the minimum distance between the $i^{\rm th}$ residential area and the $j^{\rm th}$ equity-oriented functional area. The term $\max\limits_i \{d_{\rm equ}(i,j)\} - \min\limits_i \{d_{\rm equ}(i,j)\}$ denotes the extreme difference in distance from each community to the nearest ${\rm equ}_j$ type area. The parameter $\alpha$ scales the equity reward to ensure it is within $[0,1]$. It encourages the agent to plan schools and hospitals in more equitable locations.

\subsubsection{Satisfaction.}
\begin{figure}[ht]
\centering
\includegraphics[width=1\columnwidth]{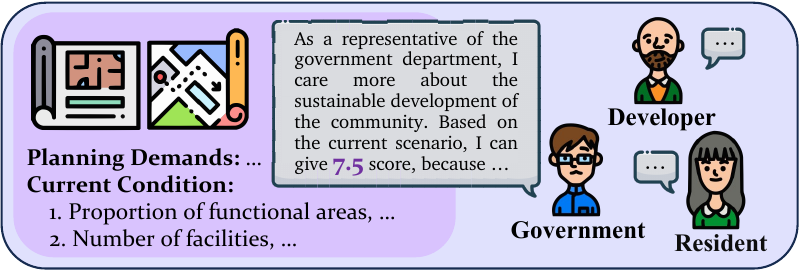}
\caption{Stakeholders are providing satisfaction scores based on planning demands and current conditions.}
\label{llm3}
\end{figure}

In recent years, participatory urban planning has become the mainstream approach in modern urban planning, involving active participation from various stakeholders~\cite{li2020collaborative}. With the rapid development of LLMs, we can leverage them for role-playing and task completion by defining specific identities~\cite{wang2023survey}. Therefore, we can have LLM-based agents act as different stakeholders, participating in the evaluation of planning schemes, which is shown in Figure~\ref{llm3}. Specifically, considering residents, developers, and government, residents may focus more on the functional areas related to daily life convenience, developers may prioritize economic benefits, and the government may aspire to build a sustainably developing community. The metric can be expressed as follows:
\begin{equation}
    r_{\rm sat} = \frac{1}{n_{\rm sta}} \sum_{i=1}^{n_{\rm sta}} {\rm Score}_i(\mathcal{P} \mid {\rm profile}_i),
\end{equation}
where $n_{\rm sta}$ is the number of stakeholders, and ${\rm Score}_i(\mathcal{P} \mid {\rm profile}_i)$ represents the satisfaction score given by a ${\rm profile}_i$ stakeholder based on the planning scheme $\mathcal{P}$.

\subsection{Intelli-Planner Training}
We employ the Proximal Policy Optimization (PPO) algorithm to train our Intelli-Planner agent. We collect training samples from the interaction between the agent and the environment using the on-policy method and update parameters in real time. Specifically:
\begin{equation}
    r_t(\theta) = \frac{\pi_{\theta}(a_t \mid s_t)}{\pi_{\theta_{\rm old}}(a_t \mid s_t)}
\end{equation}
\begin{equation}
    \hat{A}_t = Q(s_t, a_t) - v(s_t)
\end{equation}
where $\theta$ is the parameters of policy network, $r_t(\theta)$ is importance sampling, and $\hat{A}_t$ is the advantage function. Next, utilizing PPO-clip, constraints are applied in the objective function to ensure that the difference between the new and old policy is not too large. Consequently, the loss for the policy network is:
\begin{equation}
    L_{\rm actor} = \min \{ r_t(\theta) \hat{A}_t, {\rm clip}(r_t(\theta), 1-\epsilon, 1+\epsilon)\hat{A}_t \}
\end{equation}
where $\epsilon$ is a hyper-parameter, and ${\rm clip}(\cdot)$ is a clipping function that restricts the value of $r_t(\theta)$ to the range of $[1-\epsilon, 1+\epsilon]$. 
We employ Mean Squared Error (MSE) to calculate the loss between the predictions of the value network and the actual rewards:
\begin{equation}
    L_{\rm critic} = {\rm MSE}(\hat{v}_t, R_t)
\end{equation}
Here, $\hat{v}_t$ is the value estimated by the value network, and $R_t$ is the return value obtained from the reward function. Finally, we utilize the Adam optimizer~\cite{kingma2014adam} to update the network parameters.

\section{Experiments}
\subsection{Experiment Setups}
\label{exp}
\subsubsection{Study Regions.}
\begin{figure}[ht]
\centering
\includegraphics[width=1\columnwidth]{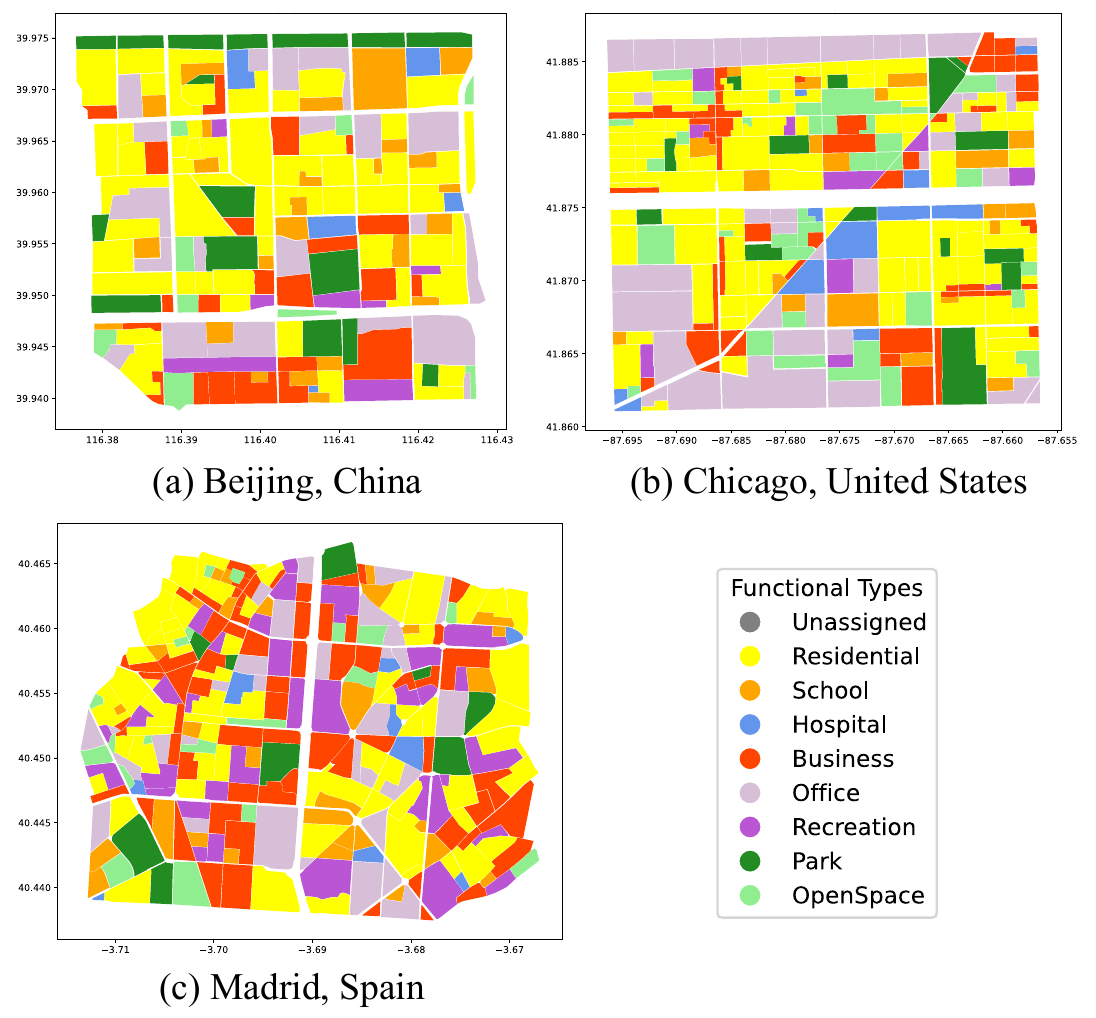}
\caption{The original functional types distribution in the planning regions of Beijing, Chicago, and Madrid.}
\label{cities}
\end{figure}

We conduct experiments in communities from three cities with different styles:
\begin{itemize}
    \item \textbf{Beijing} in China. A residential community with a high density of schools and hospitals. The community size is $4km*4km$, with the range spanning from [116$^\circ$23'E, 39$^\circ$58'N] to [116$^\circ$25'E, 39$^\circ$56'N], and a total of 34 areas for planning.
    \item \textbf{Chicago} in United States. A green and enterprise-intensive industrial community. The community size is $2.8km*3.2km$, with the range spanning from [-88$^\circ$18'W, 41$^\circ$53'N] to [-88$^\circ$20'W, 41$^\circ$51'N], and a total of 52 areas for planning.
    \item \textbf{Madrid} in Spain. A business- and entertainment-intensive tourist community. The community size is $3.2km*3.6km$, with the range spanning from [-4$^\circ$17'W, 40$^\circ$27'N] to [-4$^\circ$19'W, 40$^\circ$26'N], and a total of 57 areas for planning.
\end{itemize}
The statistics of three communities are shown in Table~\ref{tab:dataset}, and the specific functional types distribution is shown in Figure~\ref{cities}.

\begin{table}[htbp]
  \centering
  \caption{Statistics of the three communities to be planned.}
    \begin{tabular}{c|ccc}
    \toprule[1pt]
    \textbf{Cities} & \textbf{Planned/Context Area} & \textbf{\#Vacant} & \textbf{\#Plots} \\
    \midrule[0.75pt]
    \textbf{Beijing} & $2.1*2.5/4.0*4.0 (km)$ & $34$ & $158$ \\
    \textbf{Chicago} & $1.6*1.6/2.8*3.2 (km)$ & $52$ & $286$ \\
    \textbf{Madrid} & $1.7*2.2/3.2*3.6 (km)$ & $57$ & $262$ \\
    \bottomrule[1pt]
    \end{tabular}%
  \label{tab:dataset}%
\end{table}%

\begin{table*}[t]
  \centering
  \caption{Performance comparison with baselines. Obj.Score represents the objective metrics, and Sat.Score represents the satisfaction of stakeholders. The best results are \textbf{in bold}, and the second-best results are \underline{underlined}.}
  \label{tab: results}%
  \resizebox{\textwidth}{!}{
    \begin{tabular}{c|c|cccc|cc|c}
    \toprule[1pt]
    \makebox[0.12\textwidth][c]{\textbf{Cities}} & \makebox[0.122\textwidth][c]{\textbf{Methods}} & \makebox[0.07\textwidth][c]{\textbf{Service}} & \makebox[0.07\textwidth][c]{\textbf{Ecology}} & \makebox[0.07\textwidth][c]{\textbf{Economy}} & \makebox[0.07\textwidth][c]{\textbf{Equity}} & \makebox[0.07\textwidth][c]{\cellcolor[rgb]{.910, .941, .953}\textbf{Obj.Score}} & \makebox[0.07\textwidth][c]{\cellcolor[rgb]{.910, .941, .953}\textbf{Sat.Score}} & \makebox[0.12\textwidth][c]{\cellcolor[rgb]{.781, .890, .956}\textbf{Total Score}} \\
    \midrule[0.75pt]
    \multirow{7}[4]{*}{\textbf{Beijing}} & Initial &  0.4385  &  0.5843  &  0.7791  &  0.1988  & \cellcolor[rgb]{.910, .941, .953}2.0007  & \cellcolor[rgb]{.910, .941, .953}0.5060  & \cellcolor[rgb]{.781, .890, .956}2.5067  \\
          & Random &  0.4571  &  0.5042  &  0.8927  &  0.2375  & \cellcolor[rgb]{.910, .941, .953}2.0915 & \cellcolor[rgb]{.910, .941, .953}0.4940  & \cellcolor[rgb]{.781, .890, .956}2.5855 \\
          & GA &  0.4735  &  0.6670  &  0.8998  &  0.3731  & \cellcolor[rgb]{.910, .941, .953}2.4134 & \cellcolor[rgb]{.910, .941, .953}0.5694  & \cellcolor[rgb]{.781, .890, .956}2.9828  \\
          & SA &  0.4767  &  0.7423  &  0.8717  &  \textbf{0.5314} & \cellcolor[rgb]{.910, .941, .953}2.6221 & \cellcolor[rgb]{.910, .941, .953}\underline{0.7044}  & \cellcolor[rgb]{.781, .890, .956}3.3265  \\
          & PUP-MA &  \textbf{0.8363} & 0.5005 & 0.8700 & \underline{0.4691}  & \cellcolor[rgb]{.910, .941, .953}2.6759 & \cellcolor[rgb]{.910, .941, .953}0.6230  & \cellcolor[rgb]{.781, .890, .956}3.2989  \\
          & DRL-MLP & 0.4538  & \underline{0.9575}  & \textbf{1.0000} & 0.3618  & \cellcolor[rgb]{.910, .941, .953}\underline{2.7731}  & \cellcolor[rgb]{.910, .941, .953}0.6667  & \cellcolor[rgb]{.781, .890, .956}\underline{3.4398}  \\
\cmidrule{2-9}          & \textbf{Ours} &  \underline{0.4923}  &  \textbf{0.9600} &  \underline{0.9819}  &  0.4512  & \cellcolor[rgb]{.910, .941, .953}\textbf{2.8854} & \cellcolor[rgb]{.910, .941, .953}\textbf{0.7401} & \cellcolor[rgb]{.781, .890, .956}\textbf{3.6255} \\
    \midrule[0.75pt]
    \multirow{7}[4]{*}{\textbf{Chicago}} & Initial &  0.6643  &  0.7125  &  0.6642  &  0.4854  & \cellcolor[rgb]{.910, .941, .953}2.5264  & \cellcolor[rgb]{.910, .941, .953}0.6726  & \cellcolor[rgb]{.781, .890, .956}3.1990  \\
          & Random &  \textbf{0.8625} &  0.6485  &  0.6981  &  0.5288  & \cellcolor[rgb]{.910, .941, .953}2.7379 & \cellcolor[rgb]{.910, .941, .953}0.6230  & \cellcolor[rgb]{.781, .890, .956}3.3609 \\
          & GA &  0.8444  &  \underline{0.9551}  &  0.7006  &  0.6003  & \cellcolor[rgb]{.910, .941, .953}3.1004 & \cellcolor[rgb]{.910, .941, .953}0.6349  & \cellcolor[rgb]{.781, .890, .956}3.7353 \\
          & SA &  \underline{0.8593}  &  0.8880  &  0.7191  &  0.6484  & \cellcolor[rgb]{.910, .941, .953}3.1148 & \cellcolor[rgb]{.910, .941, .953}\underline{0.7321}  & \cellcolor[rgb]{.781, .890, .956}3.8469 \\
          & PUP-MA & 0.8571  & \textbf{0.9975}  & 0.7328  & \underline{0.7230}  & \cellcolor[rgb]{.910, .941, .953}3.3104  & \cellcolor[rgb]{.910, .941, .953}\underline{0.7143}  & \cellcolor[rgb]{.781, .890, .956}4.0247  \\
          & DRL-MLP & 0.8428  & 0.9420  & \underline{0.9358} & \textbf{0.7382} & \cellcolor[rgb]{.910, .941, .953}\textbf{3.4588}  & \cellcolor[rgb]{.910, .941, .953}0.6845  & \cellcolor[rgb]{.781, .890, .956}\underline{4.1433}  \\
\cmidrule{2-9}          & \textbf{Ours} &  0.8500  &  0.9470  &  \textbf{0.9713}  &  0.6891  & \cellcolor[rgb]{.910, .941, .953}\underline{3.4574}  & \cellcolor[rgb]{.910, .941, .953}\textbf{0.7540} & \cellcolor[rgb]{.781, .890, .956}\textbf{4.2114} \\
    \midrule[0.75pt]
    \multirow{7}[4]{*}{\textbf{Madrid}} & Initial &  0.8526  &  0.3690  &  \textbf{0.8501} &  0.4787  & \cellcolor[rgb]{.910, .941, .953}2.5504  & \cellcolor[rgb]{.910, .941, .953}0.6627  & \cellcolor[rgb]{.781, .890, .956}3.2131 \\
          & Random &  0.8857  &  0.9170  &  0.6644  &  0.4897  & \cellcolor[rgb]{.910, .941, .953}2.9568 & \cellcolor[rgb]{.910, .941, .953}0.6230  & \cellcolor[rgb]{.781, .890, .956}3.5798 \\
          & GA &  0.8978  &  0.9098  &  0.6318  &  0.6150  & \cellcolor[rgb]{.910, .941, .953}3.0544 & \cellcolor[rgb]{.910, .941, .953}0.6905  & \cellcolor[rgb]{.781, .890, .956}3.7449 \\
          & SA &  0.9257  &  0.9621  &  0.6817  &  \underline{0.6681} & \cellcolor[rgb]{.910, .941, .953}3.2376 & \cellcolor[rgb]{.910, .941, .953}0.6329  & \cellcolor[rgb]{.781, .890, .956}3.8705 \\
          & PUP-MA & \underline{0.9579}  & 0.8565  & 0.6448  & \textbf{0.8147} & \cellcolor[rgb]{.910, .941, .953}3.2739  & \cellcolor[rgb]{.910, .941, .953}0.7143  & \cellcolor[rgb]{.781, .890, .956}\underline{3.9882}  \\
          & DRL-MLP & \textbf{0.9600} & \textbf{1.0000}  & 0.6997  & 0.6225  & \cellcolor[rgb]{.910, .941, .953}\underline{3.2822}  & \cellcolor[rgb]{.910, .941, .953}0.6468  & \cellcolor[rgb]{.781, .890, .956}3.9290  \\
\cmidrule{2-9}          & \textbf{Ours} &  0.9400  &  \textbf{1.0000} &  \underline{0.8174}  &  0.5915  & \cellcolor[rgb]{.910, .941, .953}\textbf{3.3489} & \cellcolor[rgb]{.910, .941, .953}\textbf{0.7341} & \cellcolor[rgb]{.781, .890, .956}\textbf{4.0830} \\
    \bottomrule[1pt]
    \end{tabular}}
\end{table*}%

\begin{figure*}[ht]
\centering
\includegraphics[width=1\textwidth]{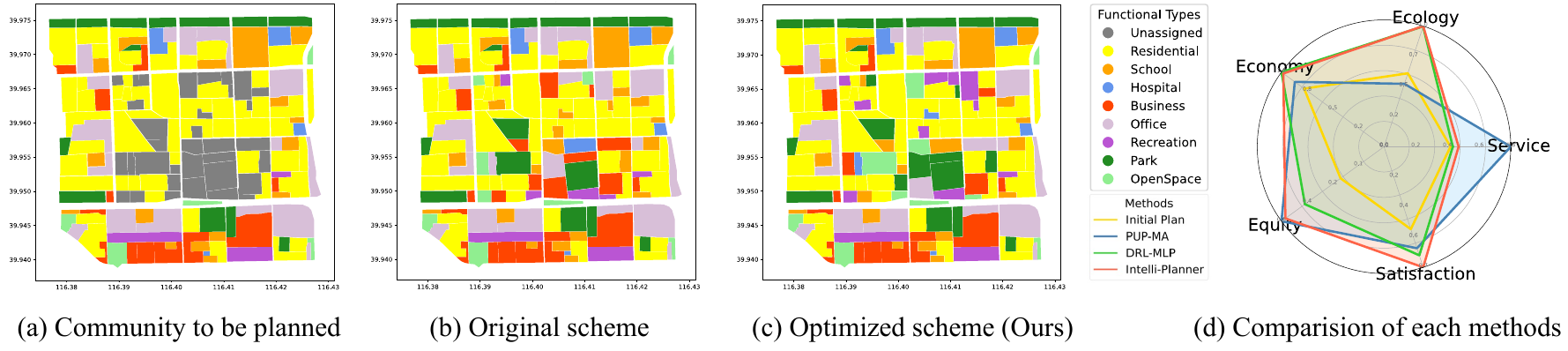}
\caption{Visualization of the original and optimized planning schemes of the Hepingli Community in Beijing.}
\label{case}
\end{figure*}

\subsubsection{Baseline Methods.}
We compare our proposed framework of Intelli-Planner with various baselines, including:
\begin{itemize}
    \item \textbf{Initial Scheme}. The actual planning schemes for the regions represent the outcomes derived from the expertise of human planners and the evolution of human living. These schemes are employed for comparison with the results obtained from various algorithms.
    \item \textbf{Random Decision}. Each area designated for planning is randomly assigned a functional type.
    \item \textbf{Genetic Algorithm (GA)}~\citep{mirjalili2019genetic}. It optimizes solutions by simulating natural selection through genetic operators such as natural selection, crossover, and mutation.
    \item \textbf{Simulated Annealing (SA)}~\citep{delahaye2019simulated}. It optimizes by mimicking the annealing process, exploring solutions iteratively, and gradually accepting less optimal choices.
    \item \textbf{PUP-MA}~\citep{zhou2024large}. In each decision step, the LLM is used to give the recommended functional type based on the planning target and the community's current planning condition.
    \item \textbf{DRL-MLP}~\citep{zheng2023spatial}. Using DRL solely to maximize the weighted sum of four objective evaluation indicators, without the knowledge enhancement in the policy network.
\end{itemize}

\subsubsection{Implementation Details.}
Our framework is implemented using a Python script, leveraging the GPT-3.5 Turbo through OpenAI API calls. The radius of the 15-minute living circle is defined as $d_c=500$ meters, and we set hyper-parameters $\alpha=-1/800$ and $\lambda=2$ for optimal performance. The network architecture for both the policy and value networks comprises layers [128, 64, 32, 8]. We adopt a learning rate of $1e-5$, and the reward decay rate $\gamma$ is set to 0.98. We give the same weight to each reward metric, adding them linearly as the total reward for learning in our framework. Training involves 5,000 episodes for each community. To ensure result reliability, we conduct 5 tests for each method and average the outcomes to obtain the final results.

\begin{figure*}[t]
\centering
\includegraphics[width=1\textwidth]{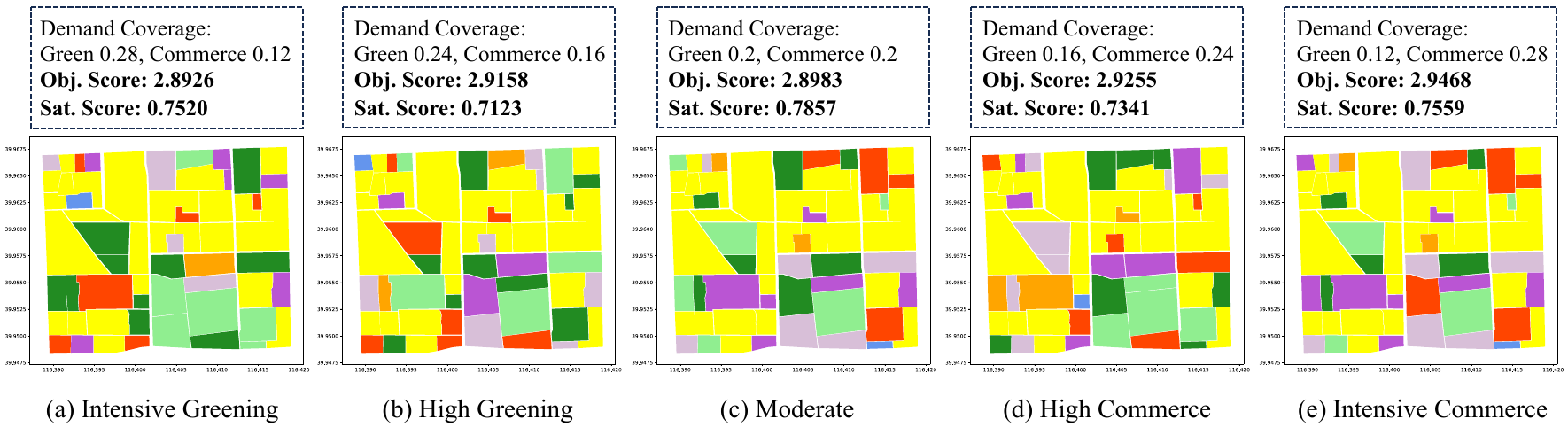}
\caption{Planning schemes of the community in Beijing under different greening and commercial demands.}
\label{style}
\end{figure*}

\subsection{Planning Results}
Table~\ref{tab: results} shows the final scores of different baselines and our framework. For each community, LLM provides planning requirements for each functional type based on planning objectives and demographic attributes.

\subsubsection{Intelli-Planner Vs. Human Planner.}
Our framework significantly outperforms initial human-designed schemes, increasing the total score by 44.63\% in Beijing, 31.65\% in Chicago, and 27.07\% in Madrid. These results show that Intelli-Planner achieves substantial improvements in overall scores and as well as shortens planning time. Similar results can be seen in the Random Decision strategy. Initial plans crafted by human experts decades ago may not align with current goals, whereas our model demonstrates flexibility by dynamically adapting to changes in needs.

\subsubsection{Intelli-Planner Vs. Optimization Algorithms.}
Additionally, we choose two heuristic approaches, genetic algorithm and simulated annealing, for comparison with our framework. Results show that our framework indeed outperforms the two heuristics on both Obj. Score and Sat. Score. Specifically, compared to the best-performing heuristics, Beijing: Obj. Score from 2.6221 to 2.8854 (+10.04\%), Sat. Score from 0.7044 to 0.7401 (+5.07\%); Chicago: Obj. Score from 3.1148 to 3.4574 (+11.00\%), Sat. Score from 0.7321 to 0.7540 (+2.99\%); and Madrid: Obj. Score from 3.2376 to 3.3489 (+3.44\%), Sat. Score from 0.6905 to 0.7341 (+6.31\%). The action space for the planning task is often large, making it challenging for traditional optimization algorithms to search for the best solution. Our Intelli-Planner can learn the mapping between community conditions and planning scheme rewards through interaction with the environment. This capability can further evolve into acquiring general planning skills.

\subsubsection{Intelli-Planner Vs. Autonomous Agents.}
Finally, we compare our framework with the most advanced LLM-based and DRL-based agents currently available. The results show that in the three communities, our proposed framework achieves the highest total score: Beijing is 3.6255 (+5.40\%), Chicago is 4.2114 (+1.64\%), and Madrid is 4.0830 (+3.92\%). DRL-MLP achieves high Obj.Score by directly optimizing objective metrics but ignores Sat.Score since it can't be quantified as a reward. Additionally, it shows significant bias in Obj.Score, resulting in large variations, whereas our approach is more balanced in four objective metrics. In contrast, PUP-MA, which relies entirely on decisions made by LLM-based agents, performs poorly in Obj.Score but achieves high Sat.Score. The introduction of Intelli-Planner effectively combines the optimization search capabilities of DRL with the role-playing strengths of LLM.

\subsection{Customized Region Planning Schemes}
In this section, we set different planning styles in the Beijing community to generate planning schemes based on various demand orientations. This stylized planning can also be applied to any other cities in our framework. Based on different proportions of greening and commercial areas, we set five different style types ranging from intensive greening to intensive commerce. Planning schemes are generated using our framework, and the results are shown in Figure~\ref{style}. We can observe that as the greening rate in the planning requirements gradually reduces and the commercial rate increases, the area of parks and open spaces in the community decreases, while the areas of commercial, office, and entertainment zones expand. Meanwhile, there are also appropriately planned schools and hospitals in the community to meet daily life needs. Despite variations in planning styles, all schemes consistently achieve high scores across diverse metrics. This underscores the effectiveness and superiority of our framework in generating customized planning schemes.

\subsection{Ablation Study}
We conduct an ablation study to analyze the LLM-only and DRL-only components separately. Table~\ref{tab: ablation} shows that LLM achieves higher Sat.Score by aligning with specific planning requirements and catering to diverse stakeholder interests, while DRL excels in meeting predefined evaluation criteria, resulting in higher Obj.Score. It demonstrates that our framework effectively combines both components to optimize satisfaction and objective evaluation.

\begin{table}[ht]
  \centering
  \caption{Ablation results for three communities using LLM-only or DRL-only in our proposed framework.}
  \resizebox{\columnwidth}{!}{
    \begin{tabular}{c|c|cc|c}
    \toprule[1pt]
    \textbf{Cities} & \textbf{Methods} & \textbf{Obj.Score} & \textbf{Sat.Score} & \textbf{Total Score} \\
    \midrule[0.75pt]
    \multirow{3}[2]{*}{\textbf{Beijing}} & LLM-only & 2.5133 & \underline{0.6806} & 3.1939 \\
          & DRL-only & \underline{2.8516} & 0.6726 & \underline{3.5242} \\
          & \textbf{Intelli-Planner} & \textbf{2.8854} & \textbf{0.7401} & \textbf{3.6255} \\
    \midrule[0.75pt]
    \multirow{3}[2]{*}{\textbf{Chicago}} & LLM-only & 3.3990 & \underline{0.7440} & 4.1430 \\
          & DRL-only & \textbf{3.4927} & 0.6925 & \underline{4.1852} \\
          & \textbf{Intelli-Planner} & \underline{3.4574} & \textbf{0.7540} & \textbf{4.2114} \\
    \midrule[0.75pt]
    \multirow{3}[2]{*}{\textbf{Madrid}} & LLM-only & 3.0215 & \underline{0.7202} & 3.7417 \\
          & DRL-only & \underline{3.3405} & 0.6925 & \underline{4.0330} \\
          & \textbf{Intelli-Planner} & \textbf{3.3489} & \textbf{0.7341} & \textbf{4.0830} \\
    \bottomrule[1pt]
    \end{tabular}}
  \label{tab: ablation}%
\end{table}%

\subsection{Analysis of LLM in Intelli-Planner}
\begin{figure}[ht]
\centering
\includegraphics[width=1\columnwidth]{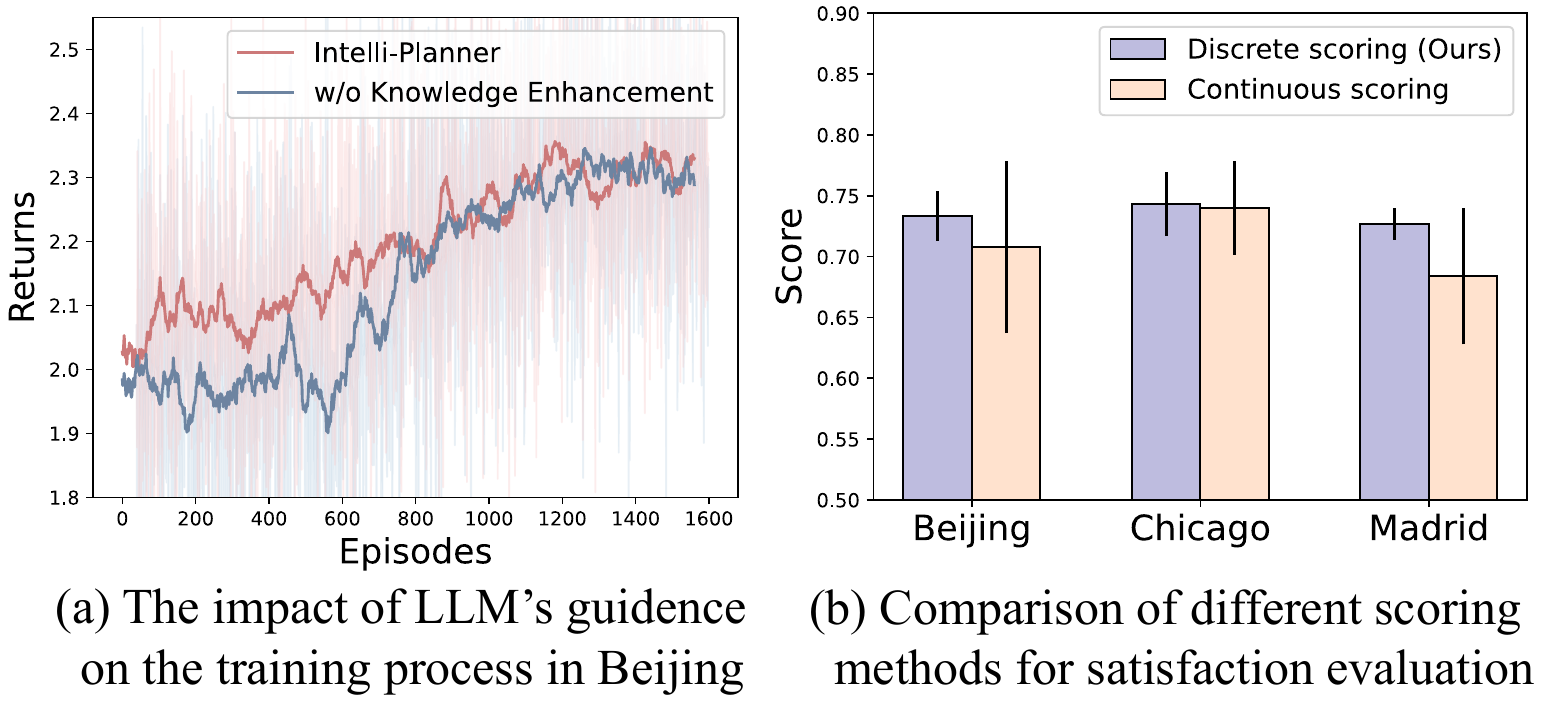}
\caption{Analysis of the LLMs at different stages.}
\label{llm}
\end{figure}

We remove the LLM-based knowledge enhancement module in the policy network and compare it with the training process of the original framework. As shown in Figure~\ref{llm}(a), this module can utilize the prior knowledge of LLM to accelerate the convergence speed in the early stage of training and improve the score of the final scheme. Then we evaluate the stability of stakeholder satisfaction ratings using LLM. Our framework employs discrete scoring, where different types of functional areas in the scheme are rated as "excellent", "good", "average", or "poor" with accompanying reasons provided. In Figure~\ref{llm}(b), discrete scoring (our method) results in a smaller variance in scores compared to continuous scoring, thereby ensuring the stability of the final results. Additionally, allowing different LLM-based stakeholders to provide detailed reasons for their scores can effectively mitigate the hallucination.

\subsection{Case Study}
We present a case study of a Beijing community in Figure~\ref{case}. Observation reveals that the scheme has achieved our established planning objectives quite well. Specifically, the community boasts high rates of greenery and open space coverage, reaching 15.62\% and 9.20\%. (The planning targets are 15\% and 10\% respectively.) While commercial and office areas are relatively scarce, accounting for only 4.87\% and 4.86\% each. (The planning targets are both 5\%.) Additionally, there are five schools and two hospitals, effectively meeting the needs for education and healthcare.

\section{Conclusion}
This paper introduces a participatory and customized urban planning framework combining LLMs and DRL. Within it, we model the planning process as a Markov Decision Process and apply LLMs in different stages. Additionally, we construct a multi-dimensional evaluation system that comprehensively assesses service, ecology, economy, equity, and satisfaction. Experimental results in three communities with different styles and needs demonstrate that our framework effectively balances objective metrics and diverse stakeholder interests, showcasing outstanding planning capabilities.

\section*{Acknowledgements}
This work was supported by the Public Computing Cloud at Renmin University of China and the Fund for Building World-Class Universities (Disciplines) at Renmin University of China.

\bibliographystyle{ACM-Reference-Format}
\bibliography{sample-base}

\appendix

\section{Predefined Planning Rules}
Based on the size of each area, we categorize the areas to be planned into two types: small and large. Among the eight types of functional zones we consider—residential, business, office, recreation, school, hospital, park, and open space—we further subdivide schools into kindergartens, primary schools, and secondary schools, and hospitals into large hospitals and regular clinics. To ensure rational planning, we stipulate that large hospitals can only be built in large areas, while clinics can only be built in small areas. Similarly, secondary schools and primary schools must be built in large areas, whereas kindergartens can be built in small areas.

\section{Differences from Existing Works}
Currently, there are two main types of work related to urban planning. The first type is based on reinforcement learning methods, while the second type is based on large language model agents. Their respective representative works are as follows:

Ref [1]. \textbf{Reinforcement Learning:} Spatial planning of urban communities via deep reinforcement learning. \url{https://www.nature.com/articles/s43588-023-00503-5}

Ref [2]. \textbf{Large Language Model:} Large language model for participatory urban planning. \url{https://arxiv.org/abs/2402.17161}

\textbf{As for Ref [1]}, although we use RL, our approach differs entirely from the technical paths:

(1) Our framework to task decomposition differs from that of Ref [1]. The planning sequence for functional types in Ref [1] is predefined, with each step focusing solely on the "where to build" question. In contrast, we base our approach on the more common steps in human planning, considering the problem as "what to build" for each area to be planned. Our approach is more flexible and can be applied to a wider variety of functional area layout types.

(2) The design details vary significantly in RL. For example, in Ref [1], the state only incorporates information from the individual parcel, while we construct state features from three levels: the parcel itself, the living circle, and the global context. Furthermore, given the different focus of our problem as mentioned above, the selection strategy for actions is also entirely different.

(3) We have developed a more comprehensive evaluation system. Regarding service, we both are inspired by the concept of a 15-minute living circle, so we propose a similar assessment. However, for ecology, Ref [1] considers the ratio of residential area in parks and open space buffers, while we evaluate the proportion of the total area of parks and open space in the entire region. Our ecological assessment is based on the idea of limited requirements of green coverage in actual planning scenarios, and can be tailored according to the planning styles, offering greater flexibility. Additionally, we have introduced other indicators to evaluate economy, equity, and satisfaction, which were not considered in Ref [1].

\textbf{As for Ref [2]}: it utilizes a large number of LLM-based agents with different roles to perform regional planning tasks. This approach may involve significant API and time costs. To address this drawback, we selected three representative roles as stakeholders to complete the decision-making tasks. This not only significantly reduces costs but also resolves the issue of potential role redundancy among agents.

\section{The Trustworthiness of LLM Agent}
we conducted a preliminary human assessment of the validity of rating stakeholder satisfaction based on large language model. We deemed this approach reasonable and feasible, which led us to decide to incorporate it into our framework. By consulting human experts (professors in the field of urban planning) and ordinary residents (individuals from diverse backgrounds) regarding their endorsement of the rationale behind LLM statements, we found that the AI agent based on GPT-3.5 was able to effectively engage in role-playing and rating tasks in urban planning. The analysis of 50 samples shows that the GPT-3.5-based agent aligns closely with human judgments. Agreement with urban planning experts is high (Cohen's $\kappa = 0.800$, 84\% exact matches), and similarly strong for ordinary residents ($\kappa = 0.821$, 86\% exact matches), indicating that the LLM can reliably approximate human assessments of stakeholder satisfaction.


\begin{table}[ht]
\centering
\caption{Alignment between LLM Agent (GPT-3.5) and Human Experts in Stakeholder Satisfaction Rating}
\label{tab:alignment}
\resizebox{\columnwidth}{!}{
\begin{tabular}{lccc}
\hline
\textbf{Stakeholder Group} & \textbf{Cohen's $\kappa$} & \textbf{Spearman $\rho$} & \textbf{Agreement Rate} \\
\hline
Urban Planning Experts & 0.800 & 0.960 & 0.84 \\
Ordinary Residents     & 0.821 & 0.969 & 0.86 \\
\hline
\textbf{Average} & \textbf{0.811} & \textbf{0.965} & \textbf{0.85} \\
\hline
\end{tabular}}
\end{table}

\section{Prompts and Case Study}
\label{app: prompts}
\subsection{Stage 1: Objectives Formulation}
Here we take the Hepingli Community in Beijing (Figure~\ref{cities}(a)) as an example, and give the specific prompt as follows. LLM will give planning objectives for each functional type based on the basic information and planning style of the community.

\noindent\textbf{Input Prompt:}
{\fontsize{8pt}{8.5pt}\selectfont
\begin{lstlisting}[language=HTML, caption={}, basicstyle=\ttfamily, breaklines=true, breakindent=9pt, backgroundcolor=\color{gray!10}, numbers=none]
Act as an experienced urban functional district planning expert and complete the task of pre-assigning functional areas based on the residential and working portraits of the area to be planned.

# INFORMATION
Now consider a 4km*4km area in Beijing:
For the portrait of the residential population: the residential area accounts for 50% of the total area.  For age distribution, 0-18 years old accounted for 10%, 18-34 years old accounted for 30%, 34-54 years old accounted for 30%, 55 years old and over 55 years old accounted for 30%. For consumption level, the low level accounted for 15%, the medium level accounted for 50%, and the high level accounted for 35%.
For the portrait of the working population: for age distribution, 0-18 years old accounted for 10%, 18-34 years old accounted for 50%, 34-54 years old accounted for 30%, 55 years old and over 55 years old accounted for 10%. For consumption level, the low level accounted for 30%, the medium level accounted for 60%, and the high level accounted for 10%.

# TARGET PLANNING STYLE
Planning style preference: a residential community with a high greening rate, low commercial rate, abundant educational and medical resources, and mainly dominated by children's schooling and elderly care.

# REQUIREMENTS
Based on the basic situation of the community, you need to carry out high-level coarse-grained planning, including:
(1) According to the total area of the region and the population of residents, the number of large hospitals and clinics to be built is given.
(2) According to the total area of the region and the proportion of minors, the number of secondary schools, primary schools, and kindergartens to be built is given.
(3) According to the consumption level and style preference of the living and working population, the degree of coverage of commercial areas, entertainment areas, and office areas is given, and the high coverage rate, medium coverage rate, and low coverage rate are selected, without giving specific values.
(4) According to the overall situation and style preference of the community, the degree of coverage of parks and open spaces is given, and the coverage rate of high coverage, medium coverage, and low coverage is selected, without giving specific values.
Just give the result. Returned in JSON format.

# ANSWER FORMAT
{
"Hospital": quantity,
"Clinic": quantity,
"Secondary": quantity,
"Primary": quantity,
"Kindergarten": quantity,
"Business": degree of coverage,
"Recreation": degree of coverage,
"Office": degree of coverage,
"Park": degree of coverage,
"Open space": degree of coverage,
}
\end{lstlisting}}

\noindent\textbf{Output:}
{\fontsize{8pt}{8.5pt}\selectfont
\begin{lstlisting}[language=HTML, caption={}, basicstyle=\ttfamily, breaklines=true, breakindent=9pt, backgroundcolor=\color{gray!10}, numbers=none]
{
  "Hospital": 2,
  "Clinic": 2,
  "Secondary": 2,
  "Primary": 3,
  "Kindergarten": 1,
  "Business": "low coverage rate",
  "Recreation": "medium coverage rate",
  "Office": "low coverage rate",
  "Park": "high coverage rate",
  "Open space": "medium coverage rate"
}
\end{lstlisting}}

Instead of directly outputting the planning coverage of each functional type, it is more reasonable for LLM to categorize the regions into "High", "Medium", and "Low" based on their respective demands. In this paper's context, "High Coverage Rate" is defined as 15\%, "Medium Coverage Rate" as 10\%, and "Low Coverage Rate" as 5\%.

Our planning objectives for the Hepingli Community are to create \textbf{a residential area with a high proportion of green spaces, limited commercial activities, abundant educational and medical resources, mainly catering to children's education and elderly care}. The outputs of the LLM reflect a high level of green cover and minimal commercial and office space, aligning well with the community's demographics. Additionally, the distribution of hospitals and schools appears to be appropriate considering the community's needs.

\subsection{Stage 2: Knowledge Enhancement}
Then we provide a detailed analysis of the prompts used to enhance the knowledge of the policy network. LLM will act as an experienced urban planner to determine the functional types of regions.

\noindent\textbf{Input Prompt:}
{\fontsize{8pt}{8.5pt}\selectfont
\begin{lstlisting}[language=HTML, caption={}, basicstyle=\ttfamily, breaklines=true, breakindent=9pt, backgroundcolor=\color{gray!10}, numbers=none]
# MISSION
Act as a professional urban planning expert and complete the urban functional area planning decision-making task based on the urban domain knowledge and the area conditions given to you.

# RULES
You will be given:
1. Planning targets.
2. Functional types around the area to be planned and the current ratio of each functional type in the whole region.
Based on the information given and your land planning expertise, you need to decide on the three most likely types of planning for the area.

# PLANNING GOALS
1. Service: Ensure that there are 5 types of service facilities around each residential area, including Business, Office, Recreation, School, and Hospital.
2. Ecology: Make the coverage area of the Park and Open Space to 0.15 and 0.1 respectively.
3. Economy: Make the valid coverage area of Business, Office, and Recreation to 0.05, 0.05, and 0.10 respectively.
4. Equity: Make the number of small schools, large schools, small hospitals, and large hospitals to 1, 5, 2, and 2 respectively. Make the distances from schools and hospitals as equitable as possible for each residential area.
5. Satisfaction: Ensure that planning decisions are made in a way that balances the interests of residents, developers, and the government.

# INSTRUCTIONS
1. Select three functional types from Open Space, Office, Business, Recreation, Hospital, School, and Park.
2. Give the answer in the required format, without the need to state reasons. If possible, output: \"[functional_type_1, functional_type_2, ...]\". If there is too little information around, you can't give a definitive answer, output: \"No\".

# SURROUNDING INFORMATION
There is a large area. Surrounding functional types information within the living circle: ['Residential', 'Business', 'Recreation', 'Park', 'OpenSpace', 'Hospital', 'School']
Current coverage area ratio of each functional type in the community: Business area is 0.0584, Recreation area is 0.0969, Office area is 0.0251, Park area is 0.1294, Openspace area is 0.0978. Current number of schools and hospitals in the community: Small school(s) is(are) 0, Large school(s) is(are) 2, Small hospital(s) is(are) 1, Large hospital(s) is(are) 1.

What's your suggestion?
\end{lstlisting}}

\noindent\textbf{Output:}
{\fontsize{8pt}{8.5pt}\selectfont
\begin{lstlisting}[language=HTML, caption={}, basicstyle=\ttfamily, breaklines=true, breakindent=9pt, backgroundcolor=\color{gray!10}, numbers=none]
[Office, Park, School]
\end{lstlisting}}

When the coverage of business and recreation areas has nearly met the target requirements (0.0584/0.05 and 0.0969/0.10 for business and recreation areas respectively), the LLM will choose to prioritize the planning of offices, parks, and schools that have not met the requirements, which shows that the LLM is able to make decisions based on its planning knowledge.

\subsection{Stage 3: Satisfaction Evaluation}
At this stage, we consider three different categories of stakeholders: urban residents, government departments, and developers. They will score the planning scheme based on their own interest needs.
\noindent\textbf{Prompts of City Resident:}
{\fontsize{8pt}{8.5pt}\selectfont
\begin{lstlisting}[language=HTML, caption={}, basicstyle=\ttfamily, breaklines=true, breakindent=9pt, backgroundcolor=\color{gray!10}, numbers=none]
Act as a city resident and score the urban planning scheme based on the stakeholders represented by the city's residents.

Scoring Criteria: Can the proportion of each functional area in the planning scheme meet the daily needs of residents, and can residents live comfortably in it?

Planning Scheme:
1. Planning targets: {PLANNING_TARGETS}
2. Detailed information: {DETAILED_INFORMATION}

The score levels from bad to good include 'poor', 'average', 'good', and 'very good', please give the score level and reasons.
\end{lstlisting}}
\noindent\textbf{Prompts of Government Department:}
{\fontsize{8pt}{8.5pt}\selectfont
\begin{lstlisting}[language=HTML, caption={}, basicstyle=\ttfamily, breaklines=true, breakindent=9pt, backgroundcolor=\color{gray!10}, numbers=none]
Act as a city's government department and score the urban planning scheme based on the stakeholders represented by the government.

Scoring Criteria: Can the proportion of each functional area in the planning scheme meet the requirements of sustainable development, and can the interests of residents and developers be taken into account at the same time?

Planning Scheme:
1. Planning targets: {PLANNING_TARGETS}
2. Detailed information: {DETAILED_INFORMATION}

The score levels from bad to good include 'poor', 'average', 'good', and 'very good', please give the score level and reasons.
\end{lstlisting}}
\noindent\textbf{Prompts of City Developer:}
{\fontsize{8pt}{8.5pt}\selectfont
\begin{lstlisting}[language=HTML, caption={}, basicstyle=\ttfamily, breaklines=true, breakindent=9pt, backgroundcolor=\color{gray!10}, numbers=none]
Act as a city developer and score the urban planning scheme based on the stakeholders represented by individual developers.

Scoring Criteria: Can the proportion of each functional area in the planning scheme meet the needs of realizing as much commercial value as possible, and will it contribute to the economic growth of the city?

Planning Scheme:
1. Planning targets: {PLANNING_TARGETS}
2. Detailed information: {DETAILED_INFORMATION}

The score levels from bad to good include 'poor', 'average', 'good', and 'very good', please give the score level and reasons.
\end{lstlisting}}
We require each stakeholder to discretely score the planning schemes for each type of functional area and provide their reasoning.

\end{document}